\documentclass[letterpaper, 10 pt, conference]{ieeeconf}

\IEEEoverridecommandlockouts                              \pdfminorversion=4

\usepackage[pdftex]{graphicx}
\usepackage{amsmath} \usepackage{amssymb}  \usepackage{amsfonts,bm}
\usepackage{siunitx}
\usepackage{xspace} 
\usepackage{booktabs, array}
\usepackage{xcolor} 
\usepackage{lib} 
\usepackage{alias} 
\usepackage{multicol} 
\usepackage{balance} 
\usepackage[labelfont=bf,size=footnotesize]{caption}
\usepackage[hang,flushmargin]{footmisc}
\usepackage[breaklinks=true,letterpaper=true,colorlinks=true,citecolor=blue,bookmarks=false]{hyperref}
\usepackage{cite}
\title{\LARGE \bf Predicting Motion Plans for Articulating Everyday Objects}\author{Arjun Gupta \; \; Max E. Shepherd \; \; Saurabh Gupta\thanks{Authors are with the University of Illinois, Urbana-Champaign. EMails:
{\footnotesize \tt \{arjung2, maxes2, saurabhg\}@illinois.edu}. Project
website with more details: \url{https://arjung128.github.io/mpao}.}}

\begin{document}
\maketitle
\thispagestyle{plain}
\pagestyle{plain}
\setlength{\textfloatsep}{5pt}

\begin{abstract}
Mobile manipulation tasks such as opening a door,
pulling open a drawer, or lifting a toilet lid require 
constrained motion of the end-effector
under environmental and task constraints. 
This,
coupled with partial information in novel environments, makes it challenging to
employ classical motion planning approaches at test time. 
Our key insight is to
cast it as a learning problem to leverage past experience of solving similar
planning problems to directly predict motion plans for mobile manipulation
tasks in novel situations at test time.
To enable this,  we develop a simulator, \simname, that simulates articulated 
objects placed in real scenes. 
We then introduce \iikname, a fast and flexible representation for 
motion plans. Finally, we learn models that use \iikname to quickly 
predict motion plans for articulating novel objects at test time. 
Experimental evaluation shows improved speed and accuracy at generating 
motion plans than pure search-based methods and pure learning methods.

 \end{abstract}

\section{Introduction}
\seclabel{intro}

As humans, when faced with everyday articulated objects as shown in
\figref{problem}, we draw upon our vast past experience to successfully
articulate them. We know to stand on the side as we pull open a oven, and where
to lean on a door to push it open. 
Very rarely do we pull open a door onto our feet, or bump into the toilet while
lifting a toilet seat. In this paper, we develop techniques that enable robots
to similarly use past experience to {\it mine} and quickly {\it predict}
strategies for articulating everyday objects in cluttered real environments.

Current work on articulating objects casts it as a motion planning problem:
given a full scan of the environment, find a robot joint trajectory that leads
the end-effector to track the trajectory that the grasp-point on the object
should follow. This suffers from both a high-sensing cost and a high-planning
cost. Building a full articulable 3D reconstruction of the environment for
collision checking and planning is expensive and time consuming. At the same
time, finding paths that conform to tight constraints on the end-effector
trajectory while not colliding with self or surrounding obstacles or the
articulating object is computationally hard. States that adhere to the given
constraint form a measure zero set among the set of all states.  This creates
issues for sampling-based motion planners which can fail to sample states that
satisfy the constraint, or must incur computation cost to project states to the
constraint manifold~\cite{kingston2018sampling, Berenson2018}.

\begin{figure}
\insertWL{2.00}{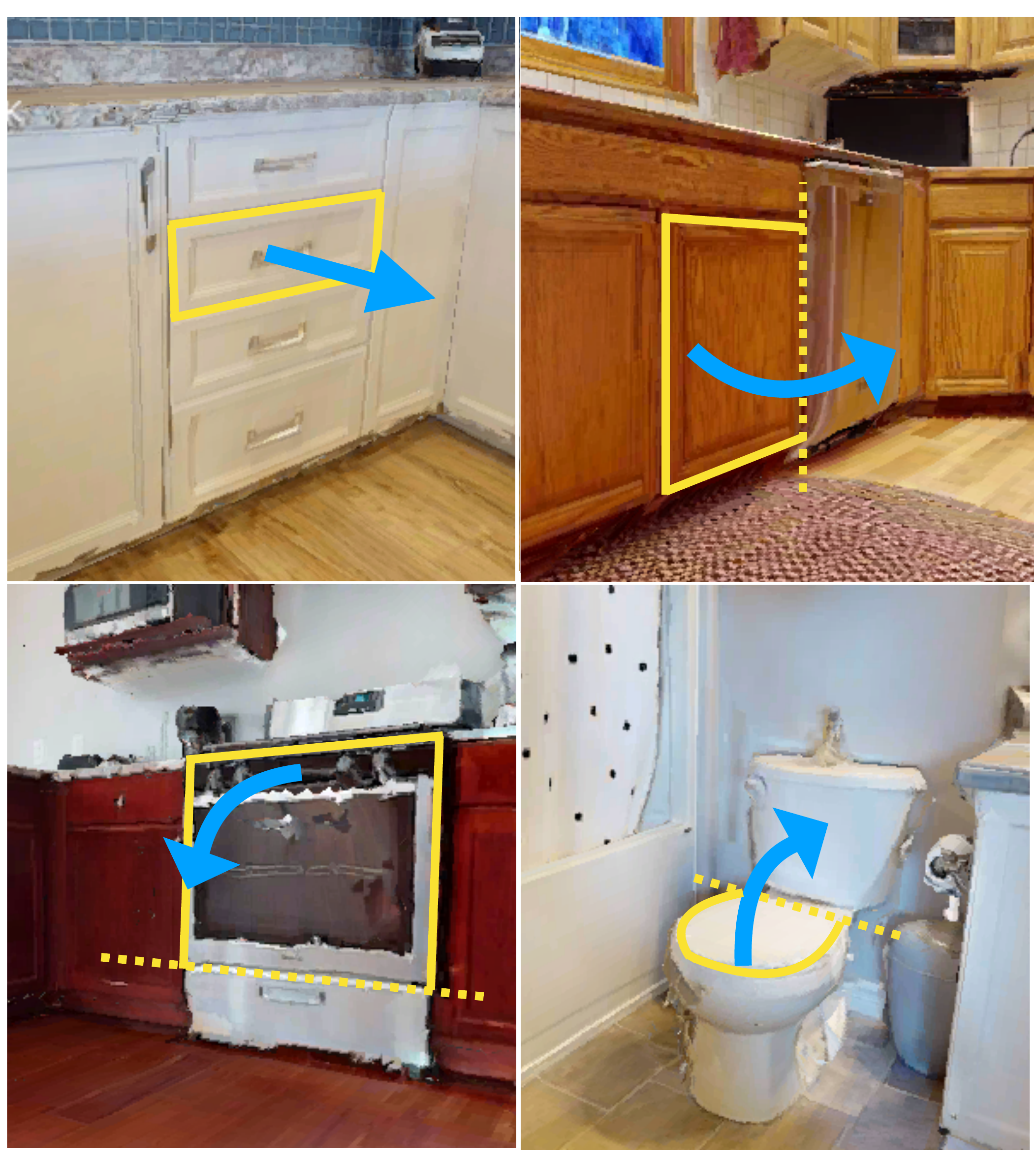} \par \vspace{2pt}
\caption{Household robots need to articulate everyday objects (\eg pull
open drawers, swing open cupboards, lift toilet seats). Such articulation
involves applying forces onto the environment while maintaining relevant
contact, such as with the drawer handle as we pull it open. This requires
reasoning about the feasibility of the entire trajectory (\ie
points along the trajectory should not just be reachable, but it must be
possible to continuously go from one point to the next). This paper
develops datasets and techniques for learning models that can predict
motion plans for such constrained motion planning problems with low
sensing and planning costs.}

\figlabel{problem}
\end{figure}
 
Rather than re-solving, from scratch, how to open a door every time we
encounter one, our proposal is to build a repertoire of strategies based on
past experience. This replaces the search in the high-dimensional motion plan
space with the much simpler problem of selecting from a small family of good
strategies, leading to gains in efficiency. Furthermore, this simpler search
can be driven by whatever observation is readily available from on-board
sensors through the use of machine learning. Our experiments demonstrate the
effectiveness of casting this as a learning problem. Given a single \rgbd
observation of an articulated object in cluttered real world scenes and
associated end-effector pose trajectory to track, we can output motion plans
that track the end-effector trajectory to within \SI{0.01}{\m} and \SI{0.01}{\radian} error with just a few
inverse kinematic calls. This, by far, outperforms the constrained motion
planning implementation for the projected state space method from the OMPL
library~\cite{kingston2018sampling, ompl} which fails to find any motion plans
with less than \SI{0.01}{\m} and \SI{0.01}{\radian} tracking error even when 
given 15 minutes of planning time.
Our impressive performance is enabled by the following three innovations.

First, we construct, \simname, a lightweight kinematic simulator for everyday
articulated objects placed in real scenes.  Crucially, this simulator
is derived from scans of {\it real-world} environments (from HM3D
dataset~\cite{ramakrishnan2021hm3d}).  This retains the appearance and the
cluttered environmental context of the articulated objects. The simulator not
only provides the experience to build the repertoire of strategies, but also
serves as the first of its kind benchmark for evaluating motion plans for
articulating objects in real environments.  \simname consists of 3758
articulated object instances across 4 articulation types (prismatic \eg
drawers, vertical hinge \eg cabinets, horizontal up-hinge \eg toilet lids,
horizontal down-hinge \eg dishwashers) across 10 object categories in 97
scenes. 

Second, rather than predicting a motion plans, that must conform to tight task
constraints and are hence hard to directly predict, we instead predict a {\it
strategy} that can be efficiently decoded into a motion plan using the
articulation geometry. 
Our decoding process consists of {\it synchronously} solving inverse kinematics
(IK) problems for end-effector waypoints sampled along the given end-effector
trajectory. This synchronization is done by warm starting IK for the
$t$\textsuperscript{th} time-step using solution from the
$(t-1)$\textsuperscript{th} time-step. We call this decoding process {
Sequential Inverse Kinematics} or \iik. By directly optimizing to reduce
end-effector pose error, \iik leads to low tracking errors. The initialization
for the first time step, $\theta_0$, serves as the strategy. Changing
$\theta_0$ changes the strategy and generates a different motion plan. 
We find that this representation, \iikname, is fast (motion plans can be quickly
decoded) and flexible (with the right $\theta_0$ it can produce
high-quality motion plans for diverse objects in diverse situations).

Not all initializations would work well for all situations. Some might not be
able to track the end-effector accurately enough, some may lead to collisions,
and others yet might violate the task constraint when joint angles are
interpolated for smooth execution. Thus, we need to find good initializations
for \iikname at test time.  Our third innovation, the use of a convolutional
neural network to predict good initializations for \iikname (or equivalently,
good strategies) from \rgb image observations, speeds up test time inference.
We train this model on a dataset of object images labeled with good
initializations, as generated using our proposed \simname simulator.  We are
able to find good solutions with only a few IK calls. This is much faster
than sampling-based planning at test time which would make tens of thousands of
IK calls to project sampled states to the constraint set. We also show that our
method can work with predicted end-effector waypoints. Collected dataset,
\simname, and code are available on the project website: \url{https://arjung128.github.io/mpao/}.

\section{Related Work}
\seclabel{related}

\noindent \textbf{Motion planning under constraints~\cite{kingston2018sampling,
Berenson2018}} has been used to tackle object articulation problems, \eg
~\cite{ruhr2012openingdoors, chitta2010planning, peterson2000high,
berenson2011task, meeussen2010autonomous, Burget2013wbmparticulated,
Vahrenkamp2013irm} among numerous other works. Researchers have tackled many
aspects: design of task-space regions for expressing constraints on
end-effectors~\cite{berenson2011task}, planning for base and arm motion
separately~\cite{meeussen2010autonomous}, considering whole-body
manipulation~\cite{Burget2013wbmparticulated}, reasoning about good locations
to position the base through inverse reachability
maps~\cite{Vahrenkamp2013irm}, and even casting it instead as a trajectory
optimization or optimal control problem~\cite{Farshidian2017brealtimeplan,
Pankert2020perceptivempc, sleiman2021unified, mittal2021articulated}.  All these approaches solve a
new object articulation problem, from scratch, every time they encounter one.
Consequently, they incur a high sensing and planning costs. Different from
these works, our interest is in techniques to leverage experience with similar
articulation problems in the past to quickly predict motion plans with low
sensing and planning cost. Online system identification
approaches~\cite{karayiannidis2016adaptive, jain2008behaviors,
niemeyer1997simple, nagatani1995experiment} that adapt plans using
feedback have also been studied. 

\noindent \textbf{Perception of articulated objects.} A body of work
\cite{wu2021vat, li2020category, Jain2020ScrewNet, zeng2020visual,
mo2021where2act, wang2019shape2motion, wang2019normalized,
ruso2009doordetection, abbatematteo2019learning, mo2019partnet,
klingbeil2010learning, anguelov2004detecting, andreopoulos2007framework} has tackled the
perception of {\it articulation} geometry for articulated objects. Given raw
sensory input (\rgb images, \rgbd images, depth
images, point clouds, or meshes) the goal
is to predict articulation parameters: \eg articulation type (prismatic \vs
hinge), segmentation of parts that independently articulate, axis of rotation /
translation, points of interaction. Researchers have 
a) investigated the use of different input modalities~\cite{
rusu2009laser, mo2021where2act, Jain2020ScrewNet, li2020category},
b) built datasets for training models~\cite{mo2019partnet},
c) designed unified output parameterizations~\cite{Jain2020ScrewNet},
and d) designed novel neural architectures and representation
~\cite{li2020category, zeng2020visual}. Researchers have also studied directly
predicting sites for interaction~\cite{mo2021where2act} and trajectories that
the robot end-effector should follow~\cite{wu2021vat} to articulate the object.
Our work is complementary, and focuses on converting articulation geometry,
possibly predicted from any of these past models, into motion plans.

\noindent \textbf{Simulators for studying object articulation} have been
challenging to build. Most past efforts use manually created {\it synthetic}
scenes: AI2-THOR~\cite{ai2thor}, Sapeins~\cite{xiang2020sapien},
ManipulaTHOR~\cite{ehsani2021manipulathor},
ThreeDWorld~\cite{gan2020threedworld}. 
Habitat 2.0~\cite{szot2021habitat} and iGibson~\cite{shen2021igibson} improve
realism by manually aligning 3D models to real scenes, but are small in size
(92 objects in 1 home and 500 objects in 15 homes respectively).
Our proposed \simname simulator is unique in its focus, studying prediction of
motion plans for everyday articulated objects, and scale, having 3758
articulated objects spread across 97 unique real world scenes. To our
knowledge, \simname is the largest dataset, to date, for the study of motion
planning performance for articulating everyday objects in everyday scenes.

\noindent \textbf{End-to-end RL approaches} can also be used to leverage prior
experience for fast execution under partial information at test
time~\cite{li2020hrl4in, xia2020relmogen, honerkamp2021learning}. However, the
large sample complexity of learning policies through RL and the small number of
environments available for training has prevented past works to show
generalization results in novel environments.  By leveraging classical
components and scaling up learning, we are able to learn models that generalize
to novel objects.

\begin{figure*}
\insertW{1.0}{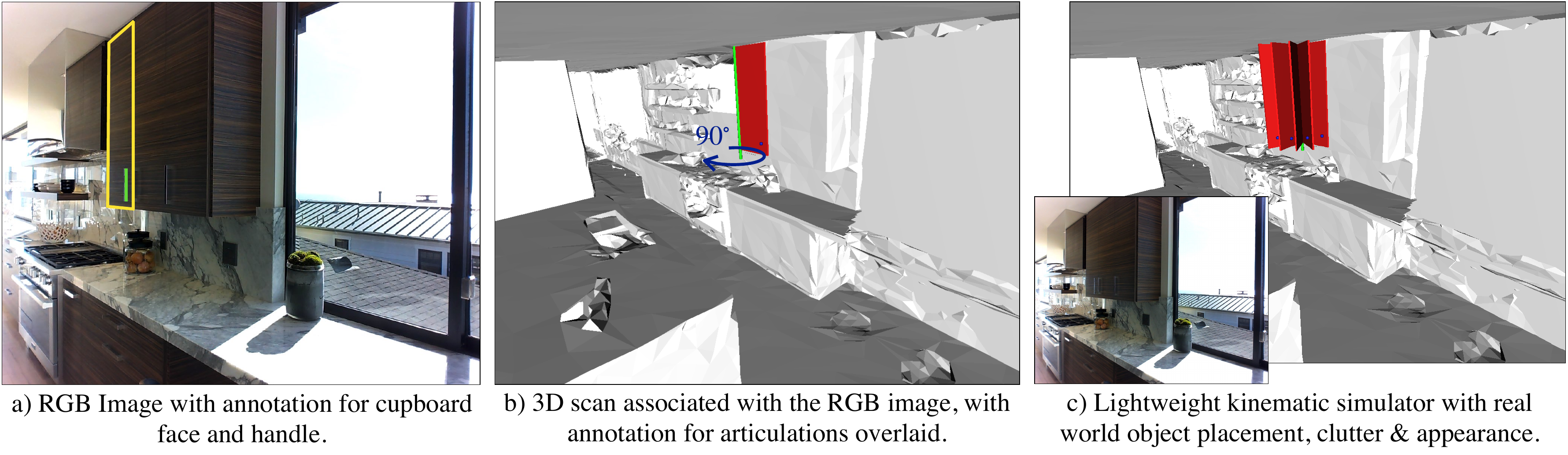}
\caption{\textbf{\simname development.} (a) We annotate \rgb images inside 3D
scans with 2D articulation geometry. (b) This is lifted to 3D using the
underlying 3D geometry. (c) As a result we get simulators that can simulate
articulated objects in realistic scenes. See \secref{dataset} for more details.} 
\figlabel{sim_door}
\end{figure*}
 
\noindent \textbf{Learning for motion planning} has been used to reduce the
runtime of motion planning algorithms: \cite{ichter2018learning,
strudel2020learning, ichter2020learned}. Strudel
\etal~\cite{strudel2020learning} learn obstacles representations for motion
planning, while Ichter \etal~\cite{ichter2020learned, ichter2018learning} use
learning to bias sampling of states for motion planners. Our use of learning is
similarly motivated, but we learn to predict low-dimensional strategies (that
can be decoded into full motion plans) for constrained motion planning problems
from visual input.

\begin{table} 
\caption{Statistics for objects and scenes in \simname, our proposed simulator
for everyday articulated objects in real scenes.}
\tablelabel{dataset}
\begin{tabular}{lrrrr}
\toprule
                                                                          &
                                                                          Train & Val & Test & Total\\
\midrule
\# Scenes                                                                 & 65    & 17  & 15   & 97\\
\# Unique Object Instances                                                & 2538  & 590 & 630  & 3758\\
\# Prismatic (\eg Drawer)                                                 & 865   & 191 & 211  & 1267\\
\# Vertical Hinge (\eg Cabinet)                                           & 1325  & 335 & 332  & 1992\\
\# Horizontal Down-hinge (\eg Oven)                                       & 146   & 40  & 40   & 226\\
\# Horizontal Up-hinge (\eg Toilet lid)                                   & 202    & 24  & 47   & 273 \\
\bottomrule
\end{tabular}
\end{table}
 
\section{\simname: A Simulator for Everyday Articulated Objects in Real Scenes}
\seclabel{dataset}
We introduce \simname, a lightweight kinematic simulator for articulated
objects placed in real scenes.
\simname is built upon the HM3D dataset~\cite{ramakrishnan2021hm3d}. HM3D
consists of 3D scans of real world environments. It offers 
both, realistic image renderings from real scenes, and access to the
underlying 3D scene geometry. \simname is made possible through 2D annotations
of articulation geometry on images, which are then lifted to 3D to allow for
a kinematic simulation of the articulated objects. To our knowledge, \simname
is the first simulator that enables a systematic large-scale study of
articulation of everyday objects in real world environments. We describe the
 steps involved in the construction of \simname.

\noindent \textbf{Annotating Articulation Geometry on Images.} The first step
is to annotate 2D articulation geometry on images. 2D articulation geometry
includes marking the extent, axis of articulation, articulation type, and
interaction locations (handles). We collect annotations in two phases.

In the first phase, we manually walk through the HM3D scenes to find kitchens
and bathrooms, and identify locations that show articulated objects.  We render
out images from different viewpoints from these locations for labeling. 

In the second phase, we use an annotation service to obtain the
necessary 2D labels. We obtain annotations for the segmentation mask for the
front face, handle locations, and articulation type (prismatic \vs left hinge
\vs right hinge \v top hinge \vs bottom hinge). See \figref{sim_door}\,(a) for an
example annotation. For most rectangular objects (\eg drawers, cupboards,
refrigerators) these three together with the underlying 3D information from the
mesh are sufficient to deduce the axis of articulation. This doesn't work for
toilets and we get additional labels for the axis of rotations (location where
the lid is attached). Toilet lids also don't have handles, we annotate and use
the lid tip as the interaction point. 

We manually verify the annotation quality after each phase and fix or reject
bad annotations. The annotation procedure is fast and cost effective (average
\$0.5 per object instance).

\noindent \textbf{Extracting 3D Articulation Geometry from 2D Annotations.} We
use the collected 2D annotations, combined with the 3D scene geometry, to
obtain a 3D simulation for each articulated object.  
For each object, we fit a plane to the points within the segmentation mask on
the depth image corresponding to the RGB image.  This gives us a 3D
representation (a 3D rectangle) for the object face that will undergo
articulation. We project the 2D handle location onto this 3D plane to obtain
the 3D handle location. Articulation parameters are obtained from this 3D
representation. We assume that the prismatic objects pull out perpendicular to
the face, and the hinged objects rotate about the corresponding edge (top,
bottom, left or right) of the 3D rectangle. As noted, toilet lids can't be
approximated as rectangles. We project the annotated 2D axis to the 3D plane.
All annotations are converted into the mesh coordinate frame using the
transformations for the camera used to render out the image.
This defines all that we need to simulate the articulating object in 3D, see
\figref{sim_door}\,(c).

\begin{figure*}
\insertW{1.0}{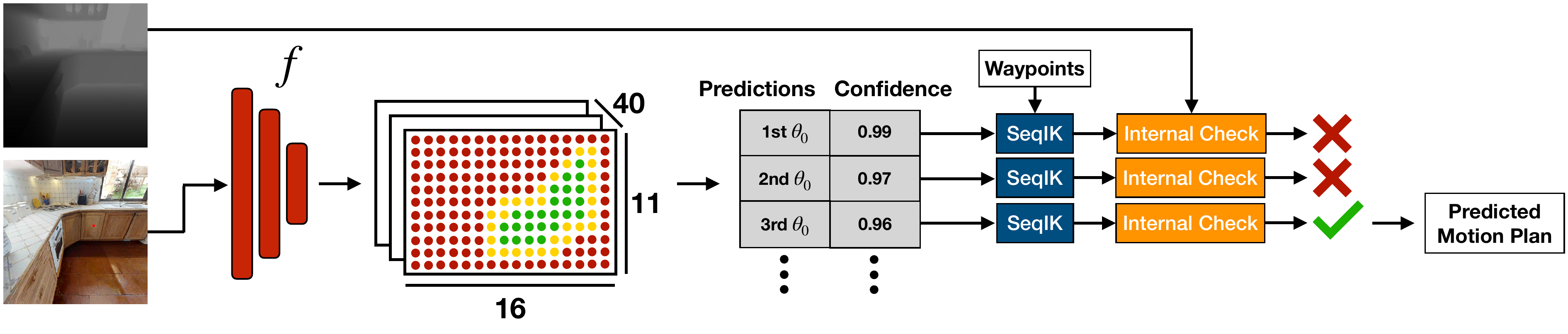}
\caption{\textbf{Overview of \fullname (Motion Plans to Articulate Objects).} 
Given an \rgbd image of the object to be articulated (denoted with a red
marker), we use a CNN to predicts good initializations for Sequential Inverse
Kinematics (\iik). \iik uses end-effector trajectories to generate motion plans
corresponding to each returned high-scoring initializations. Generated plans
are tested for deviations from the intended trajectory, and collisions using
the depth image. The first plan that succeeds these internal checks is
returned.} 

\figlabel{overview}
\end{figure*}
 
\begin{figure}
\insertWL{2.0}{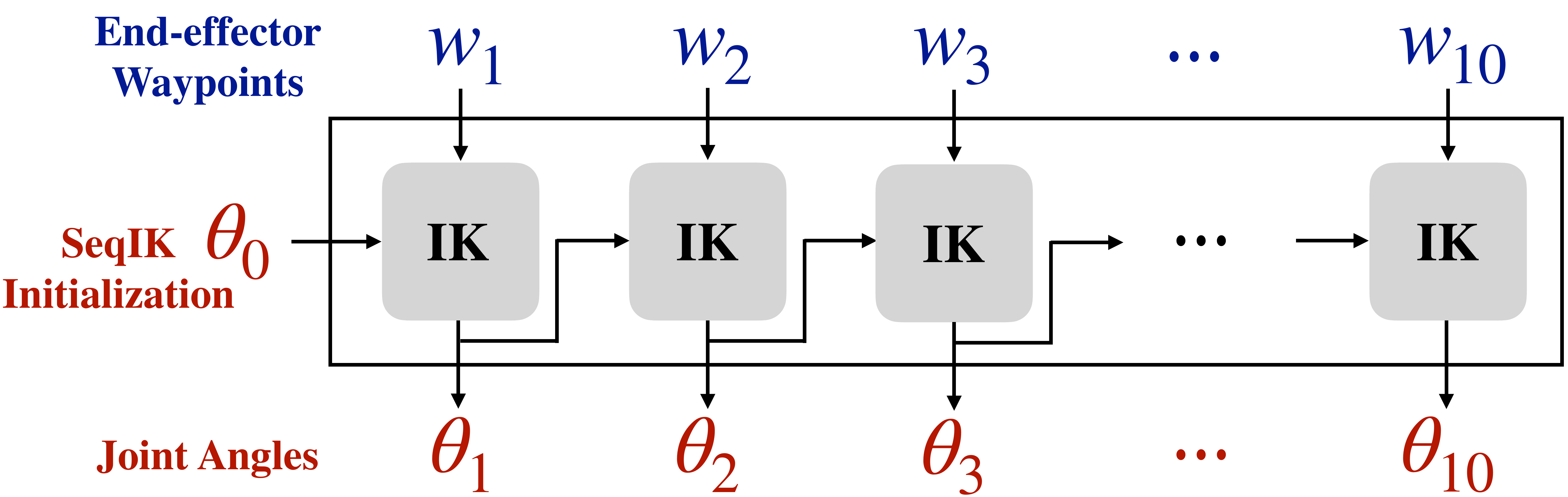} 
\caption{\textbf{Sequential Inverse Kinematics (\iik).} Given an initial joint
configuration ($\theta_0$), and a sequence of end-effector pose waypoints, \iik
uses inverse kinematics (IK) to generate configurations that achieve the given
end-effector waypoints. IK for subsequent steps is warm-started with IK
solutions from the previous time step.}
\figlabel{iik}
\end{figure}
 
\noindent \textbf{\simname Simulator.} As a result of the above two steps, we
obtain kinematic simulations for thousands of unique object instances placed in
real 3D scenes. Not only can we can simulate the object (\ie how the collision
geometry will change as the object articulates or how will the end-effector
need to move), we also have a sense of the surrounding 3D geometry of the scene
(\ie the counter below the cabinet), and can render out the \rgb appearance of
the object from multiple different views.

\tableref{dataset} shows dataset statistics. The dataset is diverse with 3758
object instances from across 97 scenes across 10 object categories and 4
articulation types.  The dataset also includes a large geometric variety \eg
cabinets high up above the counter and oven drawers very close to the ground.
This diversity, along with the fact that these objects are immersed in real
scenes makes up problem instances which have not been tackled extensively in
the literature.

In \secref{iik}, we will use \simname to design, train, and evaluate models for
predicting motion plans for articulating everyday objects.
However, we anticipate \simname will be useful for many other tasks. For
example, predicting articulation parameters or end-effector waypoints from
\rgb images, or for mining statistics about placement of articulated objects in
kitchens to build generative models for scene layout, or for building policies
for mobile manipulation.

\setlength{\tabcolsep}{8pt}
\begin{table*}
\caption{\textbf{Motion planning for articulating objects under full
information.} We measure the success rate and quality of successful plans
generated by the different motion planning methods we considered. We note that
\iikname is able to successfully generate plans quickly.  Motion planning, both
unconstrained and constrained, obtained a 0\% success rate, and hence are
omitted from the table, see \secref{mpr-results} for more details.}
\tablelabel{iik-representation-transRotComb-maxRot}
\resizebox{\linewidth}{!}{
\begin{tabular}{lS[table-format=2.1]S[table-format=1.2]S[table-format=1.4]S[table-format=4]S[table-format=5.2]}
\toprule
\bf Articulation Type                      & \multicolumn{3}{c}{\bf Performance} & \multicolumn{2}{c}{\bf Speed} \\
\cmidrule(lr){2-4} \cmidrule(lr){5-6}
& \text{Success} & \text{Translational} & \text{Rotational} & \text{Median number of} & \text{Time} \\
                                        & \text{Rate (\%)} & \text{Deviation (\si{\m)}} & \text{Deviation (\si{\radian})} & \text{initializations} & \text{(\si{s})} \\
\midrule
Prismatic (\eg Drawers)                 & 99.1       & 0.0005   & 0.0006       & 38        & 9.80     \\
Vertical Hinge (\eg Cabinets)           & 63.3       & 0.0013   & 0.0026       & 1118      & 434.12   \\
Horizontal Down-Hinge (\eg Dishwasher)  & 71.8       & 0.0029   & 0.0024       & 896       & 517.03    \\
Horizontal Up-Hinge (\eg Toilet lid)    & 44.7       & 0.0025   & 0.0028       & 1221      & 13154.12  \\
\bottomrule
\end{tabular}}
\end{table*}
 
\section{Representing and Predicting Motion Plans}
\seclabel{iik}
Given a single \rgbd image pair $[I, D]$ of an articulated object, and a
sequence of end-effector poses necessary to articulate the object $[\ldots,
w_t, \ldots]$, our next goal is to predict a motion plan, \ie sequence of joint
angles $[\ldots, \theta_t, \ldots]$ that bring the end-effector in the
necessary pose to conduct the desired articulation.  Rather than re-solving
each new problem instance from scratch using motion planning under partial
information, we pursue a machine learning approach that leverages past
experience to directly predict motion plans. A straight-forward application of
machine learning doesn't work as the predicted plans need to satisfy tight task
constraints. Instead, we use machine learning to predict a {\it strategy} 
which is decoded into a complete motion plan that adheres to the
task constraints at hand. We first describe what strategies are and
how they are decoded in \secref{plan-represent} and then describe how we
use them to predict motion plans from \rgb images in \secref{plan-predict}.

\subsection{Representing and Decoding Motion Plans}
\seclabel{plan-represent}
We represent motion plans as the initialization of a deterministic
gradient-based solver that optimizes joint angles to get the end-effector in
the desired pose. 

Our motion plan representation builds upon numerical inverse kinematics
methods~\cite{lynch2017modern}. Inverse kinematics (IK) is the process of
obtaining joint angles that get the end-effector to a given desired pose.
Starting from some initial joint angles, a numerical IK solver iteratively
updates the joint angles using the Jacobian of the forward kinematics till a
solution is found.  As we are interested in not one but a sequence of joint
angles that track the given end-effector trajectory, we {\it sequentially}
solve a sequence of inverse kinematic problems by initializing the inverse
kinematic solver for the $t$\textsuperscript{th} time-step with the solution
from the $(t-1)$\textsuperscript{th} time-step. We call this process,
Sequential Inverse Kinematics or \iik, and show a block diagram in
\figref{iik}. 

Thus, \iik can be viewed as a {\it deterministic} process that converts a
sequence of end-effector waypoints and an initial joint configuration
$\theta_0$ into a sequence of joint angles that realize the end-effector poses.
$\theta_0$ can be thought of as {\it knob} that controls the motion plans that
\iik generates. Varying $\theta_0$ varies the motion plan generated. We use
$(\text{\iik}, \theta_0)$ as our representation for {\it strategies} that
generate motion plans. Our experiments demonstrate that it is a flexible and
efficient way to generate motion plans for articulating everyday objects, and
outperforms both unconstrained and constrained motion planning approaches.

Note that \iikname, shorthand for $(\text{\iik}, \theta_0)$, may not generate 
feasible motion
plans for all inputs $\theta_0$.  Initializing from some $\theta_0$ may not get
the end-effector to where we want it to be, others might cause the end-effector
to deviate too much from the desired trajectory when interpolating between
waypoints, yet others might cause collisions with self or with the environment.
We address this issue by {\it predicting} good $\theta_0$s from the \rgb image
showing the articulated object as we describe in \secref{plan-predict}.

\subsection{Predicting Motion Plans from Images}
\seclabel{plan-predict}
Our next step is to predict good initializations $\theta_0$s for \iikname from 
\rgb images. 
As there can be more than one good $\theta_0$ for each image, we adopt
a classification approach. We work with a set of initializations $\Theta$.  We
train a function $f(I, \theta_0)$ that classifies whether or not the use of
$\theta_0$ serves as a good initialization for \iik to achieve end-effector
waypoints $[\ldots,w_t,\ldots]$ without collisions.
We provide details about the initialization set $\Theta$, function $f$,
training data, and loss function to train $f$. 

\noindent {\bf Initialization set $\Theta$} comes from the Cartesian product of
a set of robot base positions in $\mathbb{R}^3$ and a set of 10 arm
configurations. We use 704 base positions (sampled in a uniform \SI{1}{\m}
$\times$ \SI{1.5}{\m} 2D grid of base positions at a \SI{10}{\cm} resolution
at 4 different heights) and 10 arm configurations, resulting in $\Theta$ having
7040 elements.  
The 10 arm configurations are selected in a data-driven manner, we sample 20
random configurations that satisfy the joint limits, and then select the 10
which when used with \iik give us the most successes across the training set.

\noindent {\bf Function $f$} is realized through a CNN with
an ImageNet pre-trained ResNet-34 backbone~\cite{he2016deep}. We add 2 fully
connected layers on the conv5 output to produce a 7040 dimensional
representation. This is reshaped into an $80$-dimensional spatial output of size
$11 \times 16$. This is processed through another 3 convolutional layers to
produce a $(10 \cdot 4) \times 11 \times 16$ logits tensor containing $11 \times 16$
spatial output for each of the 10 arm configurations at each of the 4
heights. 

\noindent {\bf Training labels} are generated by decoding each candidate
$\theta_0$ into motion plans using \iik, and testing them for end-effector pose
deviation, self-collision, collision with the static environment, and collision
with the articulating object in our simulator from \secref{dataset}. Note
that while testing the decoded motion plans, we interpolate between consecutive
states to simulate how the plan will be executed in practice. This
process generates a binary success label for each of the 7040 candidates in
$\Theta$. This is used to supervise the logits predicted by $f$ via a binary
cross-entropy loss.

\noindent {\bf Training details}. Each articulated object instance in \simname
comes with waypoints and ground truth labels as described above. We render
multiple views for each articulated object to generate 40K images to train the
function $f$.

Our full method, {\it Motion Plans to Articulate Objects (\fullname)}, uses the
learned function $f$ to rank candidate initialization in $\Theta$. We go down
the ranked list, decode them into motion plans using \iik, and
return the first {\it feasible} plan (feasible meaning: accurately tracks the
given waypoints and also doesn't collide with self or with the geometry visible
in the depth image). See \figref{overview} for an overview.

\begin{figure}
\insertWL{1.9}{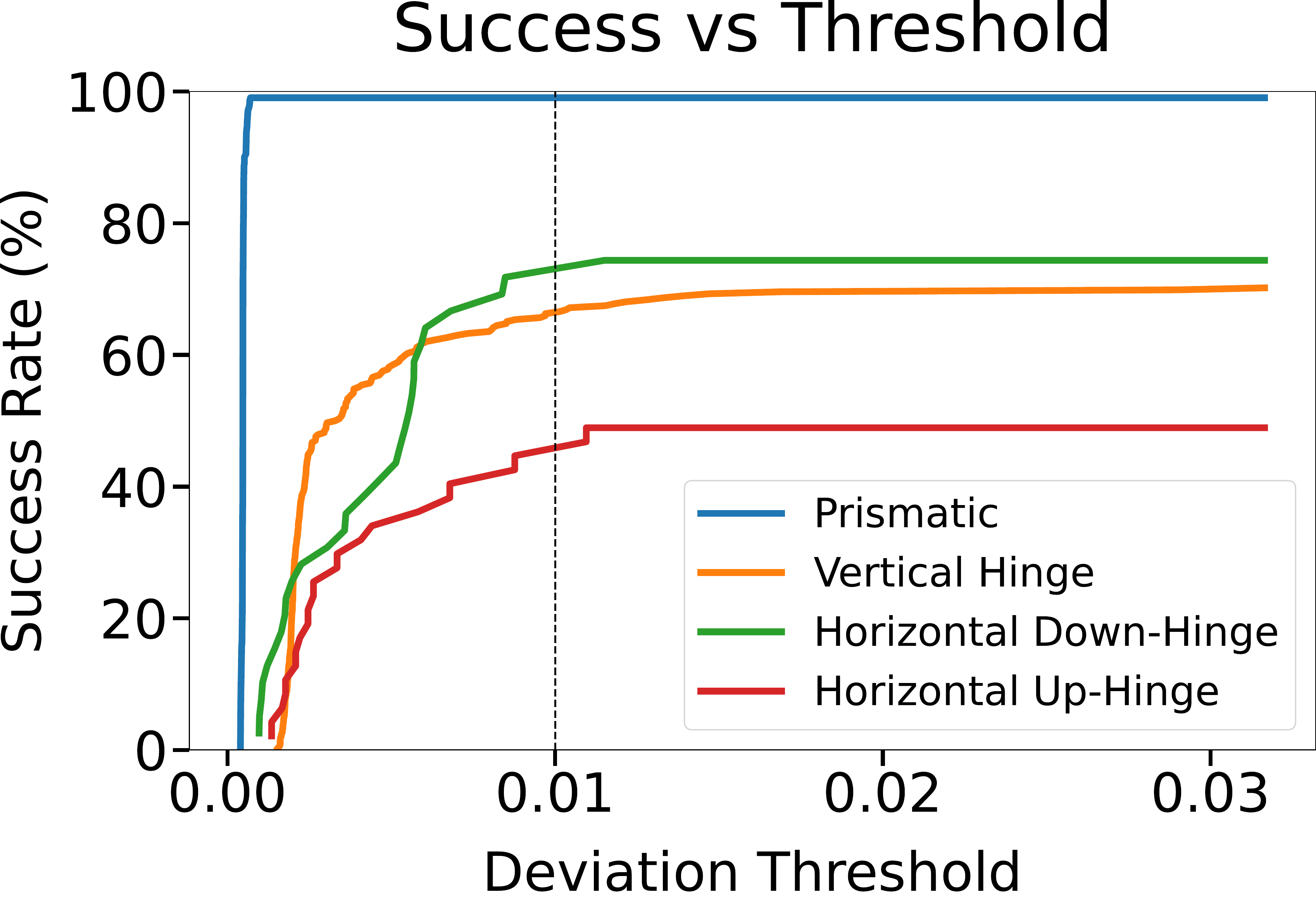} 
\caption{\textbf{Success rate as a function of deviation for motion plans 
generated by our proposed \iik.} X-axis shows maximum translational and rotational
deviation (in \si{\m} and \si{\radian} respectively), Y-axis shows the success rate for motion plans generated by \iik.
\iik generates motion plans with low deviation from desired end-effector trajectories.}
\figlabel{success_deviation}
\end{figure} 
\begin{figure*}[t]
\insertW{0.257}{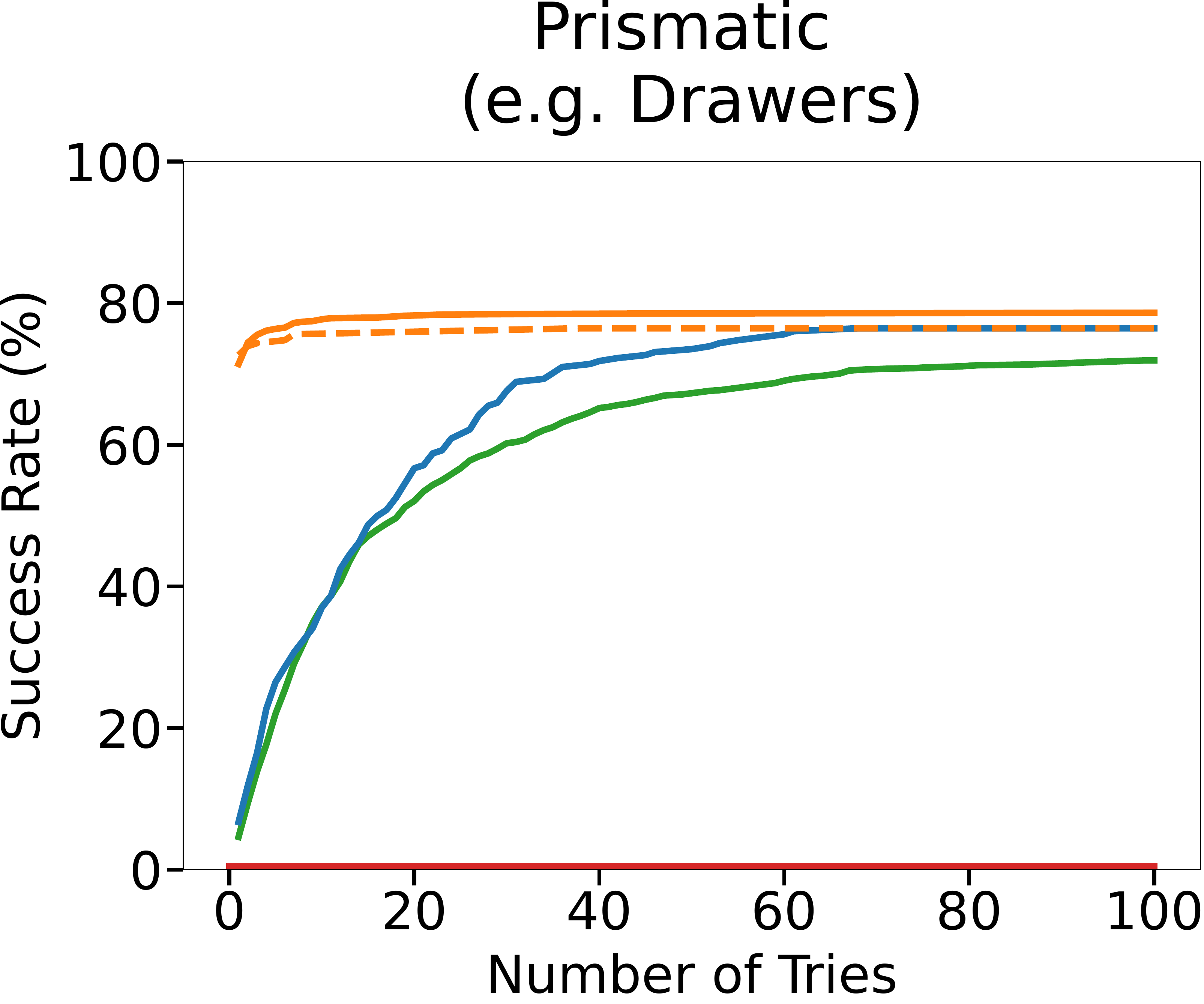}
\hfill
\insertW{0.241}{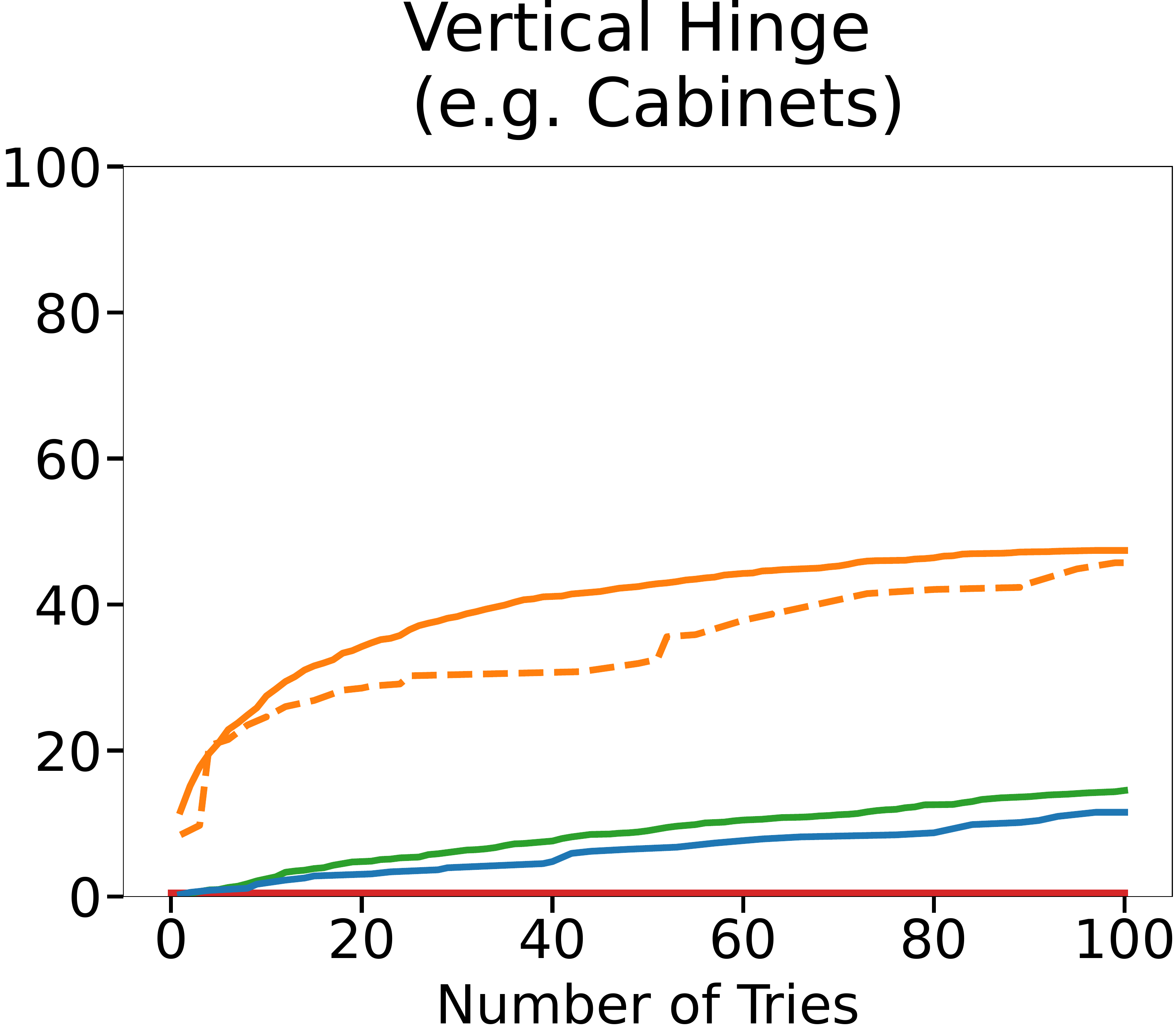}
\hfill
\insertW{0.241}{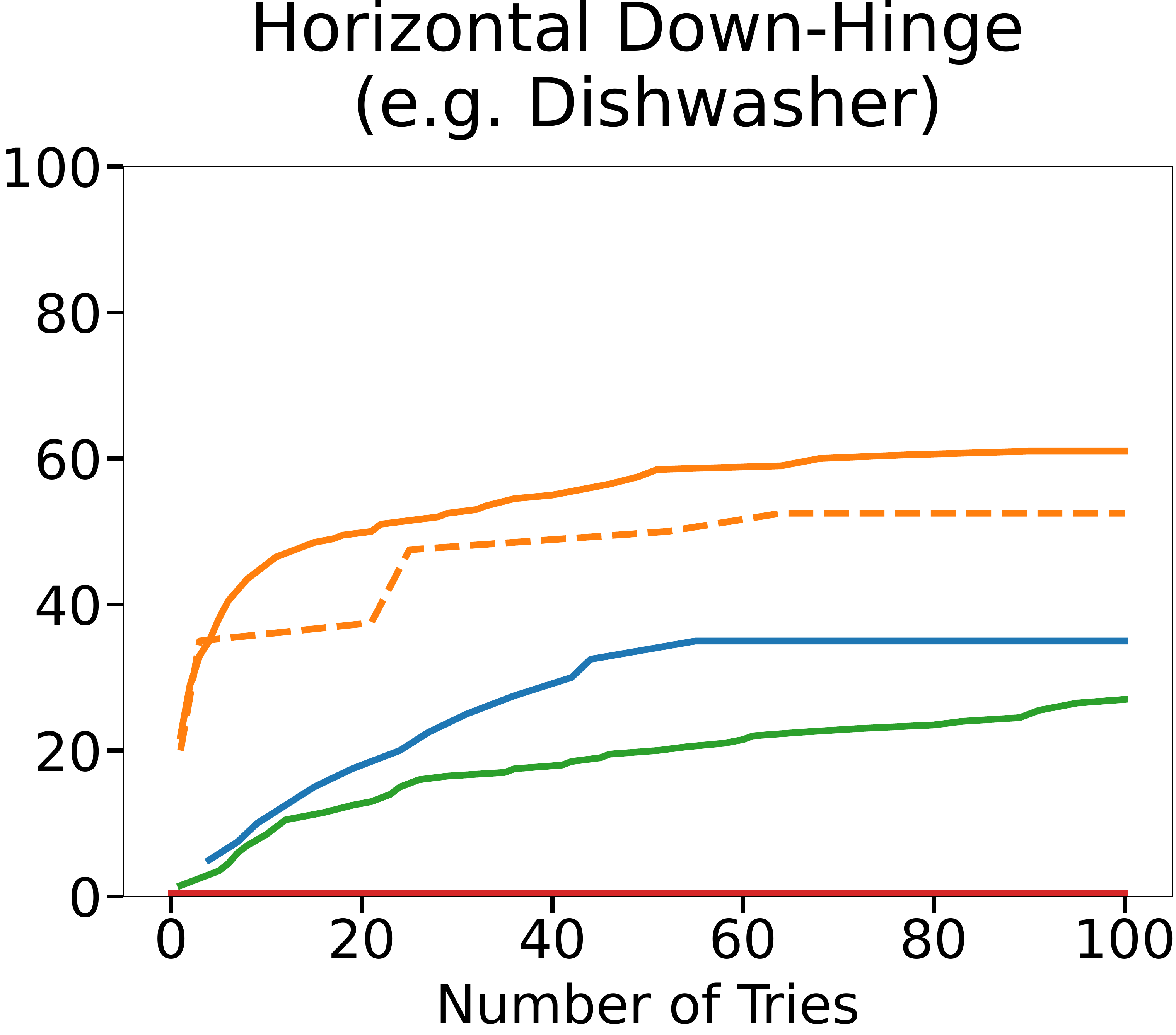}
\hfill
\insertW{0.241}{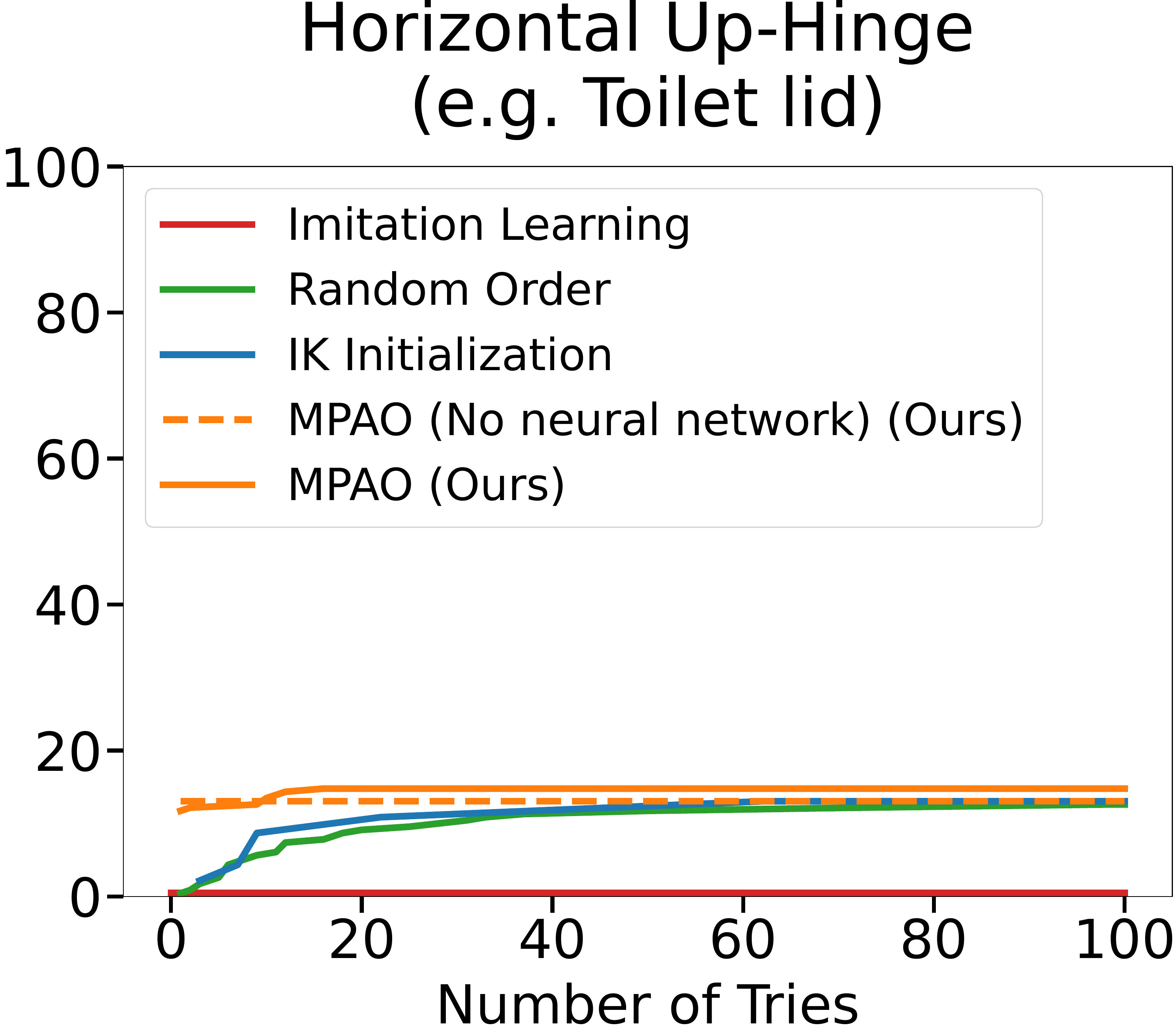}
\caption{\textbf{Motion plan prediction success rate and speed.} We show
success rate as a function of number of tries for the different articulation
types. Our method \fullname, achieves a higher success rate and generates
solutions faster than pure search or pure learning methods. 
We use \SI{0.01}{\m} translational and \SI{0.01}{\radian} rotational 
tolerance on the end-effector pose to determine success.}
\figlabel{iik-prediction}
\end{figure*}

\section{Experiments}
\seclabel{expts}
Our experiments evaluate two aspects: a) the flexibility and decoding
efficiency of our proposed motion plan representation from \secref{iik}, and b)
how effectively can we leverage \rgb images to quickly convert end-effector
poses to motion plans. For the former, we make comparisons to motion planning,
and for the latter we compare against variations that don't use the \rgb image.
We also evaluate how our method works with predicted end-effector waypoints.

\noindent {\bf Experimental Setup.} We leverage the geometry and appearance of
articulated objects in real scenes in our proposed \simname simulator 
for evaluation.
We adopt the train, val, and test splits as noted in
\tableref{dataset}. All instances from the same scene are in the same set. This
allows us to measure how well our models perform on novel held-out object
instances. We work with the 7DOF Franka Emika Panda robot. We assume that it
can take one of 4 discrete heights (\SIlist{0.25;0.5;1.0;1.5}{\m}). While we
reason about where the base should be to conduct the motion, we assume that the
base remains fixed during execution. Leveraging base motion to better
articulate objects is left to future work.

\subsection{Motion Plan Representation} 
\seclabel{mpr-results}
We evaluate the flexibility and decoding efficiency of our proposed motion
planning representation. More specifically, given a 10 time-step end-effector
trajectory and complete collision geometry of the situation, this evaluation
measures the quality of the joint angle trajectory produced by our method. We
search for a good initialization $\theta_0 \in \Theta$ for \iik and spits out
the first solution that doesn't have collisions (to self, surrounding
environment, or the articulating object) and conforms to the given tolerance in
end-effector pose.
\\ \noindent {\bf Metric.} A predicted trajectory is considered
successful if: 
a) it conveys the end-effector to the goal pose within \SI{0.01}{\m} and \SI{0.01}{\radian},
b) the resulting end-effector trajectory violates the task constraint by less
than \SI{0.01}{\m} in translation and \SI{0.01}{\radian} in rotation for each time step,
and c) it doesn't cause collisions with self, the static environment, or the
object as it articulates.  Before measuring deviations and collision-checking,
we linearly interpolate the joint angle trajectories to bring all
joint angle changes to $\leq$ \SI{0.01}{\radian}.  
\\ \noindent \textbf{Results.} 
We report the success rate and time taken by our method for different
articulation types in \tableref{iik-representation-transRotComb-maxRot}. Prismatic drawers are
easy: we can find solutions for 99.1\% of the instances to within
\SI{0.0005}{\m} and \SI{0.0006}{\radian} deviation, in as little as \SI{10}{\s} of compute while only needing to
try a median of 38 initializations. Vertical hinged and horizontal down-hinged
objects are harder: we are only able to solve 63.3\% and 71.8\% instances respectively while also
needing to sample many more initializations, taking around \SI{500}{s}.
Toilets are by far the hardest because of the tight space in bathrooms, and 
also because of the increased collision checking cost due to the non-cuboidal
toilet lid geometry. \figref{success_deviation} plots the success rate
as a function of the deviation (max translational or rotational deviation in 
\si{\m} or \si{\radian}). While we reported success rate at deviation threshold
of \SI{0.01} in \tableref{iik-representation-transRotComb-maxRot}, 
\figref{success_deviation} shows that most plans returned by \iik are much more accurate.
\\ \noindent \textbf{Comparison with other methods.} We also compared \iikname
to two other class of methods: unconstrained and constrained motion planning,
neither of which were able to find any successful solutions in a tractable
amount of time. For {\bf unconstrained motion planning}, we used
RRT-connect~\cite{kuffner2000rrt} to find a path between a start and end joint
configuration obtained using inverse kinematics. While this always found a
path, without any constraint on the intervening end-effector poses, the path
would always violate the 1-DOF constraint imposed by articulated object.  This
is not surprising as the two poses are quite far from one another. To our
surprise, even when these poses are brought close to one another, by sampling
10 way-points along the trajectory, unconstrained
motion planning would still only return solutions that would wildly swing
the end-effector around. For {\bf constrained motion planning}, we used the
projected state space method from the OMPL library~\cite{kingston2018sampling,
ompl, kingston2019exploring}. It would find motion plans that conformed to the
task constraint to some extent. However, the minimum translation deviation 
was \SI{0.02}{\m}, much
more than the tolerance level needed in our tasks, resulting again in a 0\%
success rate. We experimented with many hyper-parameter settings.
Some worked better than others, but none were able to return any plans with
less than \SI{0.02}{\m} translation deviation.

In summary, \iikname is effective at producing joint angles that conform to a
given end-effector trajectory. Finding a solution is still computationally
expensive as it requires testing many initializations. We address this using
the prediction network $f$. We evaluate it next. 

\begin{figure*}
\insertW{1.0}{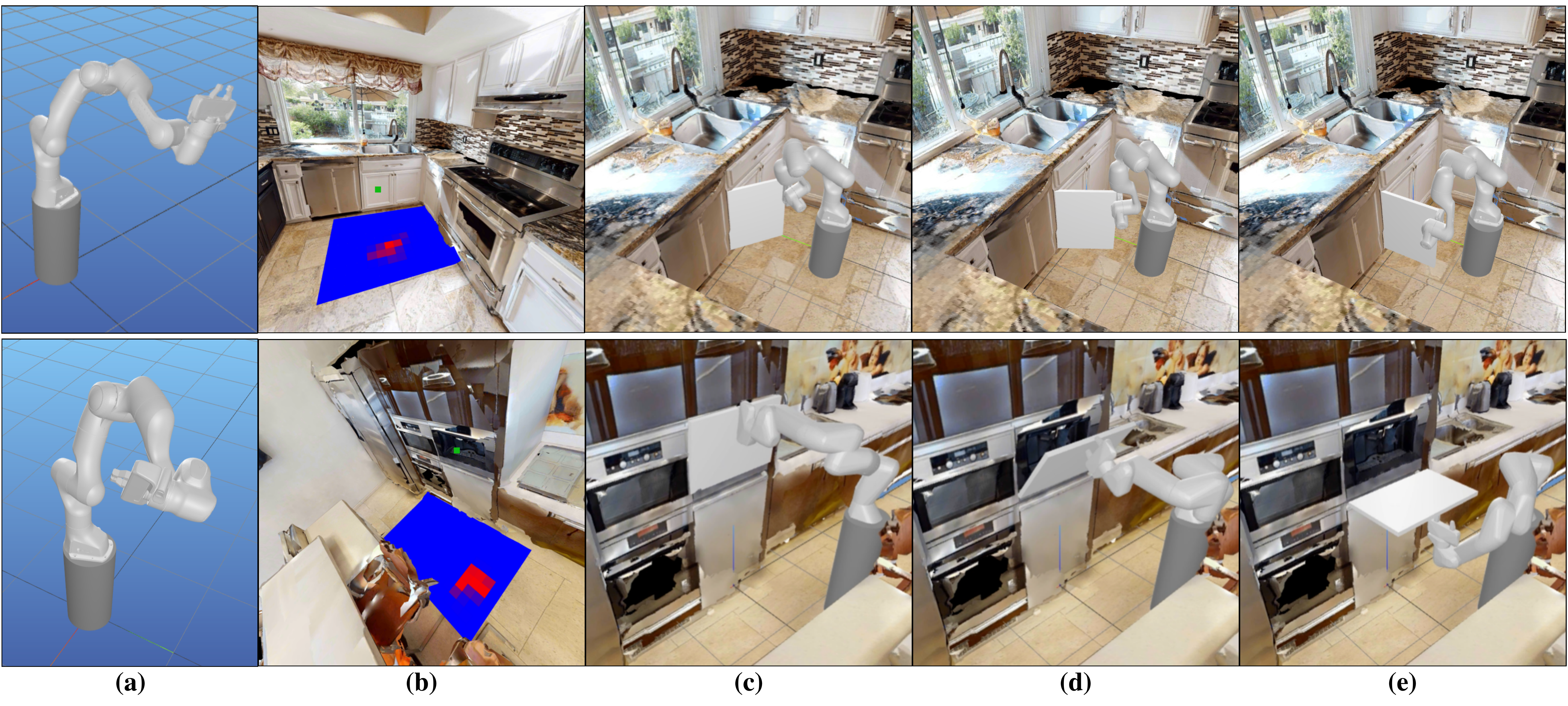}
\caption{\textbf{(a)} One of the ten arm joint configurations from $\Theta$ used for initialization.  \textbf{(b)} Example of an object from the dataset (indicated by the green marker), along with predictions for the configuration shown in (a) overlaid onto the image (warmer colors mean higher score). \textbf{(c, d, e)} Visualizations of a successful execution from one of the high-scoring locations.}

\figlabel{vis_drawer}
\end{figure*}
 
\subsection{Motion Plan Prediction with Known Waypoints}
Our next evaluation seeks to measure how quickly and accurately, we can predict
motion plans for articulated objects places in novel contexts as observed
through \rgbd images. More specifically, given an \rgbd image along with an
end-effector trajectory, we measure the success rate of predicting motion plans
as a function of planning time. As in \secref{mpr-results}, we call a predicted
motion plan successful if it reaches the goal while violating the task
constraint by less than \SI{0.01}{\m}, \SI{0.01}{\radian} and not colliding with self, the
environment, and the articulating object. While the metric is the same, the
focus of this evaluation is to assess how well methods can cope with partial
information from \rgbd observations and their speed of generating solutions.
\\ \noindent {\bf Comparisons.} 
We compare against other search schemes for finding good $\theta_0$ for \iik.
These baseline schemes employ the same overall structure as our method (\iik
decoding followed by filtering based on feasibility), but don't use any past
experience (learned model) to rank initializations. 
We consider two baselines, {\it Random Order} and {\it IK initialization}, and a
variant of our method, {\it \fullname (No neural network)}.
We also compare to a pure machine learning approach ({\it Imitation Learning})
that uses imitation learning to directly predict motion plans. We describe
these in more detail below:
\\ \textbullet\,{\it Random Order} tests whether our learned function $f$ has
learned anything meaningful about which initializations are good, and which
initializations are bad. Rather than using our learned function $f$ to sort the
set of initializations $\Theta$ for each object, we try them in a random order.
\\ \textbullet\,{\it IK initialization} uses IK to find not only the joint
angles but also the base location for the first waypoint. This way, the base
location is not restricted to our discrete grid of base positions. After this
point, \iik is used to obtain a trajectory with a fixed base position, just as
for our method.
\\ \textbullet\,{\it \fullname (No neural network) (Ours)} is a simplification
of \fullname that doesn't use the neural network $f$, but instead ranks
initializations in $\Theta$ by their success rate on the training set. Though
this doesn't use the neural network, it is still {\it data-driven} in that it
leverages experience with past constrained motion planning problems to output
plans. 
\\ \textbullet\,{\it Imitation Learning.} We also experiment with a pure
imitation learning approach that directly predicts the entire motion plan.  The
model takes in as input an RGB image (with a dot specifying which articulated
object to interact with), as well as 3D waypoints in the camera coordinate
frame. As there may be multiple correct ground truth trajectories associated
with a given input, the model outputs 100 trajectories, where each trajectory
is represented by ten joint configurations (one for each of the waypoints). We use
the Hungarian Matching algorithm~\cite{kuhn1955hungarian} to assign each
predicted trajectory the closest unique ground truth trajectory from the set of
ground truth trajectories (as also used for object detector
training~\cite{carion2020end}).  We use an $L_2$ loss on the inferred pairings
to train the model. We also jointly train a loss prediction network in order to
sort the predictions. At inference time, we employ the same internal check (as
used for all other methods) on the ranked predictions before outputting a
solution.

\noindent {\bf Results.} \figref{iik-prediction} presents the success rate
for different methods as a function of total number of solutions tried for 
novel object instances in the test set.
Across all articulation types, our method dominates pure search baselines in
success rate and speed. 
For all categories, we are able to match
baseline performance with $10\times$ fewer tries, 
and obtain more than 25\% absolute improvement in success rate for vertical and horizontal down hinges. 
Even just ranking initializations based on their
performance on the training set, \ie~{\it MPAO (No neural network)}, works
quite well. This suggests the utility of leveraging past experience for this
problem.  Our full method, MPAO, boosts performance further and is able to
effectively leverage the \rgb observation to improve the ranking among
solutions.

Imitation learning struggles for this task and yields a
$0\%$ success rate on our benchmark. We investigated this further. While
imitation learning produced the correct general motion, the motion wasn't
precise enough. When evaluated under looser criterion (max deviation of \SI{0.10}{\m}
and \SI{0.10}{\radian} \vs \SI{0.01}{\m} and \SI{0.01}{\radian} used in our benchmark), 
the imitation learning baseline obtains a non-trivial success rate (73.5\%, 2.9\%, 
15.6\% and 13.3\% respectively for the 4 categories). \fullname comfortably outperforms this even when evaluated under the tighter \SI{0.01}{\m} \SI{0.01}{\radian} criterion (success rates of 78.6\%, 47.4\%, 61.0\% and 14.8\% respectively as seen in \figref{iik-prediction}). We
also experimented with using the depth image in addition to the RGB image
for the imitation learning model, but this did not increase
performance.

These experiments together establish the effectiveness of our method at
predicting good motion plans. \figref{vis_drawer} shows an example visualization
of motion plans output by \fullname.

\subsection{Motion Plan Prediction with Unknown Waypoints}
As a proof-of-concept, we have also integrated \fullname into an overall
pipeline that doesn't require known waypoints. We experimented with drawers. We
adapt Mask RCNN~\cite{he2017mask} to detect and predict drawer faces
(segmentation mask) and handle locations (keypoints) using annotations from 
\simname. We convert them into end-effector waypoints using the depth image.
This by itself gave a median error of \SI{1.6}{\cm}. When using \fullname to
track these predicted waypoints, we are able to predict plans that solve 39\%
drawers to within \SI{0.01}{\m} translation error and 70\% to within 
\SI{0.05}{\m} translation error.

\section{Conclusion}
We pursued a learning approach that 
uses past experience to quickly predict motion plans for 
articulating objects. 
We collected a large dataset to build \simname, a simulator that enables a kinematic 
simulation of everyday objects placed in real scenes.
We designed \iikname, a fast and flexible way to represent motion 
plans under end-effector constraints, and trained neural 
network models that leverage \iikname to quickly predict plans 
for articulating novel objects.

\section*{Acknowledgements}
This material is based upon work supported by DARPA (Machine Common Sense
program), an NSF CAREER Award (IIS-2143873), the Andrew T. Yang Research and
Entrepreneurship Award, an Amazon Research Award, an NVidia Academic Hardware
Grant, and the NCSA Delta System (supported by NSF OCI 2005572 and the State of
Illinois). We thank Matthew Chang and Aditya Prakash for helpful 
feedback on experiment design and writing.

\balance

{
  \bibliographystyle{ieee_fullname.bst}
  \bibliography{biblioShort, refs, doors, refs2}  }

\end{document}